\documentclass[onecolumn, 11pt]{IEEEtran}

\usepackage[margin=1in]{geometry}
\usepackage{graphicx}
\usepackage[tight,footnotesize]{subfigure}
\usepackage{amsmath}
\usepackage{amssymb}
\usepackage{fancyhdr,url}
\usepackage{algorithmic,algorithm,cite,endnotes,balance}

\usepackage{array}

\newfloat{algo}{thp}{lop}
\floatname{algo}{Algorithm}

\newtheorem{proposition}{Proposition}[section]

\floatstyle{ruled}

\begin{document}

\title{Estimating Rigid Transformation Between Two Range Maps Using Expectation Maximization Algorithm}

\author{Shuqing~Zeng\textsuperscript{$\dag$}
\thanks{\textsuperscript{$\dag$}Department of Computer Science and Engineering, Michigan State University, East Lansing, MI 48824, e-mail:zengshuq@msu.edu.}}
\maketitle

\begin{abstract}
We address the problem of estimating a rigid transformation between two point sets, which is a key module for target tracking system using Light Detection And Ranging (LiDAR). A fast implementation of Expectation-maximization (EM) algorithm is presented whose complexity is $O(N)$ with $N$ the number of scan points.
\end{abstract}

% \linenumbers
\section{INTRODUCTION}
Rigid registration of two sets of points sampled from a surface has been widely investigated (e.g., \cite{Besl92,Granger2002,Sharp2002,Jagannathan2005,Makadia2006}) in computer vision literature. Generally, these methods are designed to tackle range maps with dense points for non-realtime applications.

In \cite{Censi2008,Olson2009} scans are matched using iterative closest line (ICL), a variant of ``normal-distance'' form of ICP algorithm~\cite{Besl92} originally proposed in computer vision community by \cite{Chen1992}. However, the convergence of this approach is sensitive to errors in normal direction estimations \cite{Stewart2006}.

\begin{figure}[h]
\centering
{\includegraphics[width=2.8in]{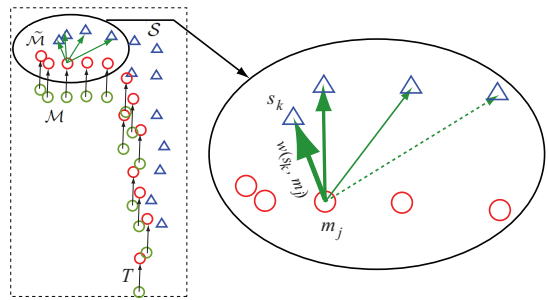}}
\caption{Illustration of the proposed algorithm. }
\label{fig:shape_registration_idea}
\end{figure}
Fig.~\ref{fig:shape_registration_idea} illustrates the concept. The light green circles denote the contour of a target $\mathcal{M}$. The red circles are the projection of $\mathcal{M}$ under a rigid transformation $T$, denoted as $\tilde{\mathcal{M}}$. Let $\mathcal{S}$ be the current range image shown as upper triangles. We propose an Expectation-maximization (EM) algorithm~\cite{Mclachain08,Granger2002} to find the rigid transformation such that the projected range image best matches the current image. Each point $m_j$ in $\tilde{\mathcal{M}}$ is treated as the center of a {\em Parzen window}. There is an edge between $s_k \in \mathcal{S}$ and $m_j$ if $s_k$ lies in the window. The weight of the edge $(s_k, m_j)$ is based on the proximity between the two vertices. The larger weight of the edge, the thicker the line is shown, and the more force that pulls the corresponding the point $m_j$ to $s_k$ through $T$.

This document describes a fast implementation of expectation maximization (EM) algorithm~\cite{Mclachain08} to locally match between $\mathcal{M}$ and $\mathcal{S}$. By exploiting the sparsity of the locally matching matrix, this implementation scales linearly with the number of points.

\section{ALGORITHM DERIVATION} \label{SC:Algorithm}
\label{SC:EM}
This section is devoted to the problem of how to estimate the rigid transformation $T$ using (EM) algorithm, giving scan map $\mathcal{S}$ and a contour model $\mathcal{M}$.

We constructed a bipartite graph $B=(\mathcal{M}, \mathcal{S}, E_B)$ between the vertex set $\mathcal{M}$ to $\mathcal{S}$ with $E_B$ the set of edges. Let $m \in \mathcal{M}$ and $s\in \mathcal{S}$. An edge exists between the points $s$ and $m$ if and only if $\| s-m\|<W $ with $W$ a distance threshold. By $\mathcal{N}(s) \equiv \{m \mid (s,m) \in E_B\}$ we denote the neighborhood of $s$.

Scan points are indexed using a lookup hash-table with $W/2$ resolution. Find the points $m$ near a point $s$ within the radius $W$ involving searching through all the three-by-three neighbor grid of the cell containing $s$. Since hash table is used, and $|\mathcal{N}(s)|$ is bounded, construction graph $B$ is an $O(N)$ operation with $N$ the number of points in a scan.

Let $s_j \in \mathcal{S}$ be one of the $n_{\mathcal{S}}$ scan points, and $m_k \in \mathcal{M}$ be one of the $n_{\mathcal{M}}$ points from the model. We denote $T$ a rigid transformation from the model to the new scan frame, with the parameter vector $\mathbf{y}$. If $s_j$ is the measure of $m_k$ (i.e.,  $(s_j, m_k) \in \mathcal{B}$) with a known noise model, we write the density function as $p(s_j\mid m_k, \mathbf{y}) = p(s_j \mid T(m_k, \mathbf{y}))$. In case of an additive and centered Gaussian noise of precision matrix $\Gamma$, $p(s_j\mid m_k, T) = c \exp(-\frac{\|s_j-T(m_k,\mathbf{y})\|_\Gamma^2}{2})$ where the Mahalanobis norm is defined as $\|x\|_\Gamma^2 \equiv x^T\Gamma x$.

We use the binary matrix $A$ to represent the correspondence between $s_j$ and $m_k$. The entry $A_{jk}=1$ if $s_j$ matches $m_k$ and 0 otherwise. Assume each scan point $s_j$ corresponds to at most one model point. We have
\[
\Sigma_k A_{jk}=\left \{\begin{array}{l} 1 ~~\mbox{If } \mathcal{N}(s_j) \neq \emptyset \\0~~\mbox{Otherwise.} \end{array}\right.
\]
for all scan point index $j$.

For the above equation, we note that for the case $\mathcal{N}(s_j) = \emptyset$, $s_j$ is an outlier, and the correspondence $s_j$ to $m_k$ can be treated as a {\em categorical distribution}.
In order to apply EM procedure we use a random matching matrix $\mathcal{A}$ with each element a binary random variable. Each eligible matching matrix $A$ has a probability $p(A)\equiv p(\mathcal{A}=A)$. One can verify that $\bar{A}_{jk} = \mbox{E}\{\mathcal{A}_{jk}\}= P(\mathcal{A}_{jk}=1)$, and the following constraint holds
\[
\Sigma_k \bar{A}_{jk}=\left \{\begin{array}{l} 1 ~~\mbox{If } \mathcal{N}(s_j) \neq \emptyset \\0~~\mbox{Otherwise.}\end{array}\right.
\]

Considering the distribution of $\mathcal{A}_j$, the $j$-th row of the $\mathcal{A}$, which is the distribution of assigning the scan point $s_j$ to the model point $m_k$, i.e.,
\[
p(\mathcal{A}_j) = \prod_{m_k \in \mathcal{N}(s_j)}  (\bar{A}_{jk})^{\mathcal{A}_{jk}}
\]
Assuming the scan points are independent, we can write
\begin{equation}
p(\mathcal{A}) = \prod_{s_j \in \mathcal{S}} \prod_{m_k \in \mathcal{N}(s_j)}  (\bar{A}_{jk})^{\mathcal{A}_{jk}} = \prod_{(s_j,m_k)\in E_{\mathcal{B}}} (\bar{A}_{jk})^{\mathcal{A}_{jk}} \label{eq:matching_pdf}
\end{equation}
An example of $p(\mathcal{A})$ is the noninformative prior probability of the matches: a probability distribution that a given scan point is a measure of a given model point without knowing measurement information:
\[
\bar{A}_{jk}=\pi_{jk}=\left\{\begin{array}{l} \frac{1}{|\mathcal{N}(s_j)|}~~\mbox{If } \mathcal{N}(s_j) \neq \emptyset \\ 0~~\mbox{Otherwise.} \end{array}\right.
\]

The joint probability of the scan point $s_j$ and the corresponding assignment $\mathcal{A}_{j}$ can be expressed as
\[
p(s_j, \mathcal{A}_{j}\mid \mathcal{M}, \mathbf{y}) = \prod_{m_k \in \mathcal{N}(s_j)} (\pi_{jk} p(s_j \mid m_k, \mathbf{y}))^{\mathcal{A}_{jk}}
\]
Providing that the scan points are conditionally independent, the overall joint probability is the product of the each row of $A$:
\begin{equation}
p(\mathcal{S}, \mathcal{A}\mid \mathcal{M}, \mathbf{y}) = \prod_{(s_j, m_k) \in E_{\mathcal{B}}} (\pi_{jk} p(s_j \mid m_k, \mathbf{y}))^{\mathcal{A}_{jk}} \label{eq:jointprob}
\end{equation}
and the logarithm of marginal distribution can be written as
\begin{equation}
\mbox{ML}(T) = \log p(\mathcal{S} \mid \mathcal{M}, \mathbf{y}) =\log \left(\sum_{\mathcal{A}} p(\mathcal{S}, \mathcal{A}\mid \mathcal{M}, \mathbf{y}) \right)
\label{eq:ML}
\end{equation}
Unfortunately, Eq.~\eqref{eq:ML} has no closed-form solution and no robust and efficient algorithm to directly minimize it with respect to the parameter $\mathbf{y}$. Noticing that Eq.~\eqref{eq:ML} only involves the logarithm of a sum, we can treat the matching matrix $\mathcal{A}$ as latent variables and apply the EM algorithm to iteratively estimate $\mathbf{y}$. Assuming after $n$-th iteration, the current estimate for $\mathbf{y}$ is given by $\mathbf{y}_n$, we can compute an updated estimate $T$ such that $\mbox{ML}(T)$ is monotonically increasing, i.e.,
\[
\Delta(\mathbf{y}\mid \mathbf{y}_n) = \mbox{ML}(\mathbf{y}) - \mbox{ML}(\mathbf{y}_n) > 0
\]
Namely, we want to maximize the difference $\Delta(\mathbf{y}\mid \mathbf{y}_n)$.

Now we are ready to state two propositions whose proofs are relegated to Appendix.
\begin{proposition}
\[\Delta(\mathbf{y}\mid \mathbf{y}_n) = \mbox{E}_{\mathcal{A}\mid \mathcal{S}, \mathcal{M}, \mathbf{y}_n} \{ \log \left( {p(\mathcal{S}, \mathcal{A}\mid \mathcal{M},\mathbf{y})}\right) \}\]
\end{proposition}
\begin{proposition}
Given the transformation estimate $\mathbf{y}_n$, scan points $\mathcal{S}$ and model points $\mathcal{M}$, the posterior of the matching matrix $A$ can be written as
\begin{equation}
p(\mathcal{A}\mid \mathcal{S}, \mathcal{M}, \mathbf{y}_n) = \prod_{j,k:(s_j, m_k) \in E_{\mathcal{B}}} (\hat{A}_{jk})^{\mathcal{A}_{jk}} \label{eq:matching_posterior}
\end{equation}
where
\begin{equation}
\mbox{E}{\{\mathcal{A}\}_{jk}}=\hat{A}_{jk} = \left\{ \begin{array}{l} \frac{\pi_{jk} p(s_j \mid m_k, \mathbf{y}_n)} { \sum_k \pi_{jk} p(s_j \mid m_k, \mathbf{y}_n)} ~~\text{If } \mathcal{N}(s_j) \neq \emptyset  \\ 0 ~~\text{Otherwise.} \end{array}\right. \label{eq:posterior_update}
\end{equation}
\end{proposition}

Therefore, we have the following EM algorithm to compute $\mathbf{y}$ that maximizes the likelihood defined in Eq.~\eqref{eq:ML}. We assume there exists an edge in the graph $\mathcal{B}$ between $s_j$ and $m_k$ in the following derivation.
\begin{itemize}
\item \textbf{E-step:} Given the previous estimate $T_n$, we update $\hat{A}_{jk}$ using Eq.~\eqref{eq:posterior_update}. The conditional expectation is computed as
{\small \begin{align}
\Delta(\mathbf{y}\mid \mathbf{y}_n) &= \mbox{E}_{\mathcal{A}\mid \mathcal{S}, \mathcal{M}, \mathbf{y}_n} \left\{ \log p(\mathcal{S}, \mathcal{A}\mid \mathcal{M}, \mathbf{y}) \right\} \nonumber \\
&= \mbox{E} \left\{ \log \left( \prod_{j,k} \pi_{jk} p(s_j \mid m_k, \mathbf{y})^{\mathcal{A}_{jk}} \right) \right\} \nonumber\\
  &= \sum_{j,k} \mbox{E}\{\mathcal{A}_{jk}\}(\log p(s_j \mid m_k, \mathbf{y}) + \log \pi_{jk}) \nonumber \\
  & = \sum_{j,k} \hat{A}_{jk} \| s_j - T(m_k, \mathbf{y})\|_\Gamma^2  + \mbox{const.} \label{eq:E_step}
\end{align}}
where $\mbox{E}$ is $\mbox{E}_{\mathcal{A}\mid \mathcal{S}, \mathcal{M}, \mathbf{y}_n}$ in short, and const. is the terms irrelevant to $\mathbf{y}$.

\item \textbf{M-step:} Compute $\mathbf{y}$ to maximize the least-squares expression in Eq.~\eqref{eq:E_step}.
\end{itemize}

The above EM procedure is repeated until the model is converged, i.e., the
difference of log-likelihood between two iterations $\Delta(\mathbf{y}\mid \mathbf{y}_n)$ is less than a small number.
The complexity of the above computation for a target in each iteration is $O(|E_{\mathcal{B}}|)$. Since the number of neighbors for $s_j$ is bounded, the complexity is reduced to $O(|\mathcal{S}|)$.  Since experimental result shows that only 4-5 epochs are needed for EM iteration to converge. Consequently, the overall complexity for all of the tracked objects is $O(N)$ with $N$ the number of scan points.

The following proposition shows how to compute the covariance matrix for the transformation parameters $\mathbf{y}$.
\begin{proposition}
Given $\mathbf{y}$, the covariance matrix $\mathbf{R}$ is
\begin{equation}
\mathbf{R}  =  \frac{1}{n_P}\sum_{(s_j, m_k) \in E_{\mathcal{B}}} \hat{A}_{jk}(s_j - T(m_k, \mathbf{y}))(s_j - T(m_k, \mathbf{y}))^T \label{eq:covariance}
\end{equation}
where $n_P$ is the number of the nonzero rows of the matrix $\hat{A}$.
\end{proposition}

\section{Proof of Propositions}

\subsection{Proof of Proposition 2.1}
{
\begin{align}
&\Delta(\mathbf{y}\mid \mathbf{y}_n) = \mbox{ML}(\mathbf{y}) - \mbox{ML}(\mathbf{y}_n) \nonumber \\
    & = \log \left(\sum_{\mathcal{A}} p(\mathcal{S}, \mathcal{A}|\mathcal{M}, \mathbf{y}) \right)  - \log \left( p(\mathcal{S}\mid \mathcal{M}, \mathbf{y}_n) \right) \nonumber\\
    & = \log \left(\sum_{\mathcal{A}} p(\mathcal{S}| \mathcal{A},\mathcal{M}, \mathbf{y}) p(\mathcal{A} | \mathcal{M}, \mathbf{y}) \right) \nonumber\\
    &~~~~ - \log  p(\mathcal{S}| \mathcal{M}, \mathbf{y}_n) \nonumber\\
    & = \log \left(\sum_{\mathcal{A}} p(\mathcal{A}| \mathcal{S}, \mathcal{M}, \mathbf{y}_n) \frac{p(\mathcal{S}| \mathcal{A},\mathcal{M}, \mathbf{y}) p(\mathcal{A} | \mathcal{M}, \mathbf{y})}{p(\mathcal{A}| \mathcal{S}, \mathcal{M}, \mathbf{y}_n)} \right) \nonumber \\
    &~~~~ - \log  p(\mathcal{S}| \mathcal{M}, \mathbf{y}_n) \nonumber\\
    & \geq \sum_{\mathcal{A}} p(\mathcal{A}| \mathcal{S}, \mathcal{M}, \mathbf{y}_n) \log \left(\frac{p(\mathcal{S}| \mathcal{A},\mathcal{M}, \mathbf{y}) p(\mathcal{A} | \mathcal{M}, \mathbf{y})}{p(\mathcal{A}| \mathcal{S}, \mathcal{M}, \mathbf{y}_n)}\right)\nonumber\\
    &~~~~ - \log  p(\mathcal{S}| \mathcal{M}, \mathbf{y}_n) \label{eq:jansen_inequ}\\
    & = \sum_{\mathcal{A}} p(\mathcal{A}| \mathcal{S}, \mathcal{M}, \mathbf{y}_n) \log \left(\frac{p(\mathcal{S}| \mathcal{A},\mathcal{M}, \mathbf{y}) p(\mathcal{A} | \mathcal{M}, \mathbf{y})}{p(\mathcal{A}| \mathcal{S}, \mathcal{M}, \mathbf{y}_n) p(\mathcal{S}| \mathcal{M}, \mathbf{y}_n)}\right) \nonumber
\end{align}}where Jansen's inequality and convexity of logarithm function are applied in deriving Eq.~\eqref{eq:jansen_inequ}. Since we are maximizing $\Delta(\mathbf{y}| \mathbf{y}_n)$ with respect to $\mathbf{y}$, we can drop terms that are irrelevant to $\mathbf{y}$, thus
{ \begin{align}
\Delta(\mathbf{y}| \mathbf{y}_n) &= \sum_{\mathcal{A}} p(\mathcal{A}| \mathcal{S}, \mathcal{M}, \mathbf{y}_n) \log \left({p(\mathcal{S}| \mathcal{A},\mathcal{M}, \mathbf{y}) p(\mathcal{A} | \mathcal{M}, \mathbf{y})}\right) \nonumber\\
& = \sum_{\mathcal{A}} p(\mathcal{A}| \mathcal{S}, \mathcal{M}, \mathbf{y}_n) \log \left( \frac{p(\mathcal{S}, \mathcal{A}, \mathbf{y}| \mathcal{M})}{p(\mathcal{A}, \mathbf{y} | \mathcal{M})}\frac{p(\mathcal{A}, \mathbf{y} | \mathcal{M})}{p(\mathbf{y} | \mathcal{M})}\right) \nonumber \\
& = \sum_{\mathcal{A}} p(\mathcal{A}| \mathcal{S}, \mathcal{M}, \mathbf{y}_n) \log \left( \frac{p(\mathcal{S}, \mathcal{A}, \mathbf{y}| \mathcal{M})}{p(\mathbf{y} | \mathcal{M})}\right) \nonumber\\
& = \sum_{\mathcal{A}} p(\mathcal{A}| \mathcal{S}, \mathcal{M}, \mathbf{y}_n) \log \left( {p(\mathcal{S}, \mathcal{A}| \mathcal{M}, \mathbf{y})}\right) \nonumber\\
& = \mbox{E}_{\mathcal{A}| \mathcal{S}, \mathcal{M}, \mathbf{y}_n} \{ \log \left( {p(\mathcal{S}, \mathcal{A}| \mathcal{M},\mathbf{y})}\right) \}
\end{align}}

\subsection{Proof of Proposition 2.2}
If $\mathcal{N}(s_j) \neq \emptyset$, the marginal PDF of the $j$-th row of $A$ is $p(s_j | \mathcal{M}, \mathbf{y}) = \sum_{k} \pi_{jk} p(s_j | m_k, \mathbf{y})$. We assume there exists an edge in the graph $\mathcal{B}$ between the scan and model points $s_j$ and $m_k$, and scan points are independent each other. One can verify that
\[
p(\mathcal{S}| \mathcal{M}, \mathbf{y}_n) = \prod_j \left(\sum_k \pi_{jk} p(s_j | m_k, \mathbf{y}_n)\right)
\]
Using Bayesian theorem, we have
{ \[
\begin{split}
p(\mathcal{A}| \mathcal{S}, \mathcal{M}, \mathbf{y}_n)   &= \frac{p(\mathcal{S}, \mathcal{A} | \mathcal{M}, \mathbf{y}_n)}{p(\mathcal{S} | \mathcal{M}, \mathbf{y}_n)} \\
& = \frac{\prod_{j,k} \left(\pi_{jk} p(s_j | m_k, \mathbf{y}_n)\right)^{\mathcal{A}_{jk}}} {\prod_j \left(\sum_k \pi_{jk} p(s_j | m_k, \mathbf{y}_n)\right)} \\
& = \frac{\prod_{j,k} \left(\pi_{jk} p(s_j | m_k, \mathbf{y}_n)\right)^{\mathcal{A}_{jk}}} {\prod_{j,k} \left(\sum_k \pi_{jk} p(s_j | m_k, \mathbf{y}_n)\right)^{\mathcal{A}_{jk}}}\\
& = \prod_{j,k} \left( \frac{\pi_{jk} p(s_j | m_k, \mathbf{y}_n)} { \sum_k \pi_{jk} p(s_j | m_k, \mathbf{y}_n)} \right)^{\mathcal{A}_{jk}}
\end{split}
\]}
Comparing with Eq.~\eqref{eq:matching_posterior}, the equation Eq.~\eqref{eq:posterior_update} holds.

\subsection{Proof of Proposition 2.3}
We treat the precision matrix $\Gamma$ as the uncertainty of unknown transformation parameter $\mathbf{y}$. We use a maximum likelihood approach, which amounts to minimizing Eq.~\eqref{eq:E_step} with respect to $\Gamma$ given a transformation and a set of matches with probabilities:
\begin{align}
     &\frac{\partial}{\partial \Gamma} \Delta(\mathbf{y}| \mathbf{y}_n)  \nonumber \\
     &=\frac{\partial}{\partial \Gamma} \sum_{(s_j, m_k) \in E_{\mathcal{B}}} \hat{A}_{jk} \left(\frac{\| s_j - T(m_k,\mathbf{y})\|_\Gamma^2}{2}  +  \log |\Gamma|^{-\frac{1}{2}}\right)\nonumber \\
     & = \frac{1}{2}\sum_{(s_j, m_k) \in E_{\mathcal{B}}} {\hat{A}_{jk} (s_j - T(m_k,\mathbf{y}))(s_j - T(m_k,\mathbf{y}))^T} \nonumber\\
     & - \frac{n_P}{2} \Gamma^{-1} = 0 \nonumber
     \end{align}
where $n_P$ is the number of nonzero rows of the matrix $\hat{A}$.
Thereby the covariance matrix $\mathbf{R}$ is computed as
\begin{align}
\mathbf{R} &=\Gamma^{-1} \nonumber\\
&= \frac{1}{n_P} \sum_{(s_j, m_k) \in E_{\mathcal{B}}} \hat{A}_{jk}(s_j - T(m_k))(s_j - T(m_k))^T \nonumber
\end{align}

\bibliographystyle{ieee}
\bibliography{gm_eci}

\end{document}